\documentclass{article}
\usepackage{graphicx} 
\usepackage{vilyastyle}
\usepackage{hyperref}
\usepackage{booktabs}
\usepackage{multirow}
\usepackage[backend=biber,style=numeric, natbib=true, sorting=none]{biblatex}
\title{Accurate structural modeling of chemically diverse molecular interfaces with Vilya-2}
\author{Vilya Research}

\begin{document}

\maketitle
\thispagestyle{titlepage}

\begin{abstract}
Structure-prediction networks built on co-evolutionary statistics have transformed protein-based drug discovery, yet their accuracy does not extend to peptide therapeutics---an increasingly important modality defined by non-canonical residues, macrocyclization, and complex topologies. We introduce Vilya-2, a diffusion transformer that extends the all-atom representation of Vilya-1 from modeling individual molecules to modeling their interactions with protein targets. This all-atom representation enables transfer learning between different molecular types, and delivers highly accurate structural modeling of peptides across sizes, classes, and compositions bound to therapeutically relevant targets. By generating diverse structural ensembles and ranking them with calibrated confidence, Vilya-2 recovers 59.1\% of peptide interfaces to sub-2 \AA~backbone RMSD, far exceeding the performance of a representative co-folding model even when that model is given the bound receptor as a template. In addition, Vilya-2 is state-of-the-art at small-molecule docking, and generalizes to novel protein-small molecule complexes unlike those seen in training. It also generalizes to modeling molecular conformations of diverse macrocycles and disulfide-stapled miniproteins several-fold larger than any molecule seen in training. Finally, Vilya-2 can be used as a foundation model, and fine-tuned to enrich for active compounds in hit-to-lead campaigns. By unifying predictive accuracy with broad generalizability across chemical space, Vilya-2 is the structure-prediction oracle that \textit{de novo} peptide design pipelines require---establishing the all-atom approach as a general foundation for the design and evaluation of \textit{de novo} of peptide therapeutics. 

To accelerate method development for this modality, we release Riptides, a curated, chemically diverse benchmark of protein-peptide complexes: \href{http://github.com/VilyaPublic/Riptides}{github.com/VilyaPublic/Riptides}.

\end{abstract}

\section{Introduction}
Peptide-based therapeutics are an increasingly important modality with widespread use in diabetes (insulin, semaglutide), recent approvals in psoriasis (Icotyde) \cite{gooderham2025lb1142}, tantalizing efficacy in cardiometabolic disease (Lipfendra) \cite{navar2026placebo}, and potentially transformative therapies in oncology (zolucatetide, AUBE00, daraxonrasib, CID-078) \cite{klempner2025fog001, kage2026kras, wolpin2026daraxonrasib, shapiro2026orally, dowlati2026cid078}. As a modality, peptides offer the promise of antibody-like potencies - with their ability to augment the biology of complicated protein-protein interactions - in a molecule format that is amenable to once-daily oral dosing \cite{wang2022therapeutic}. However, peptide drugs are often deceptively complex in structure: they can be composed of amino acid building blocks that extend far beyond the canonical 20 amino acids and arranged in more diverse topologies than linear peptides. Historically, most peptide drugs have been derived from naturally occurring peptides, and very few result from truly \textit{de novo} design. While experimental-based discovery platforms like mRNA- or phage-display can be used to identify novel peptides as starting points for drug discovery efforts, advancing the initial hits is arduous as they often lack desired drug-like features such as cell-membrane permeability, metabolic stability, and target specificity.

\begin{figure}[htbp]
  \centering
  \includegraphics[width=1.0\linewidth]{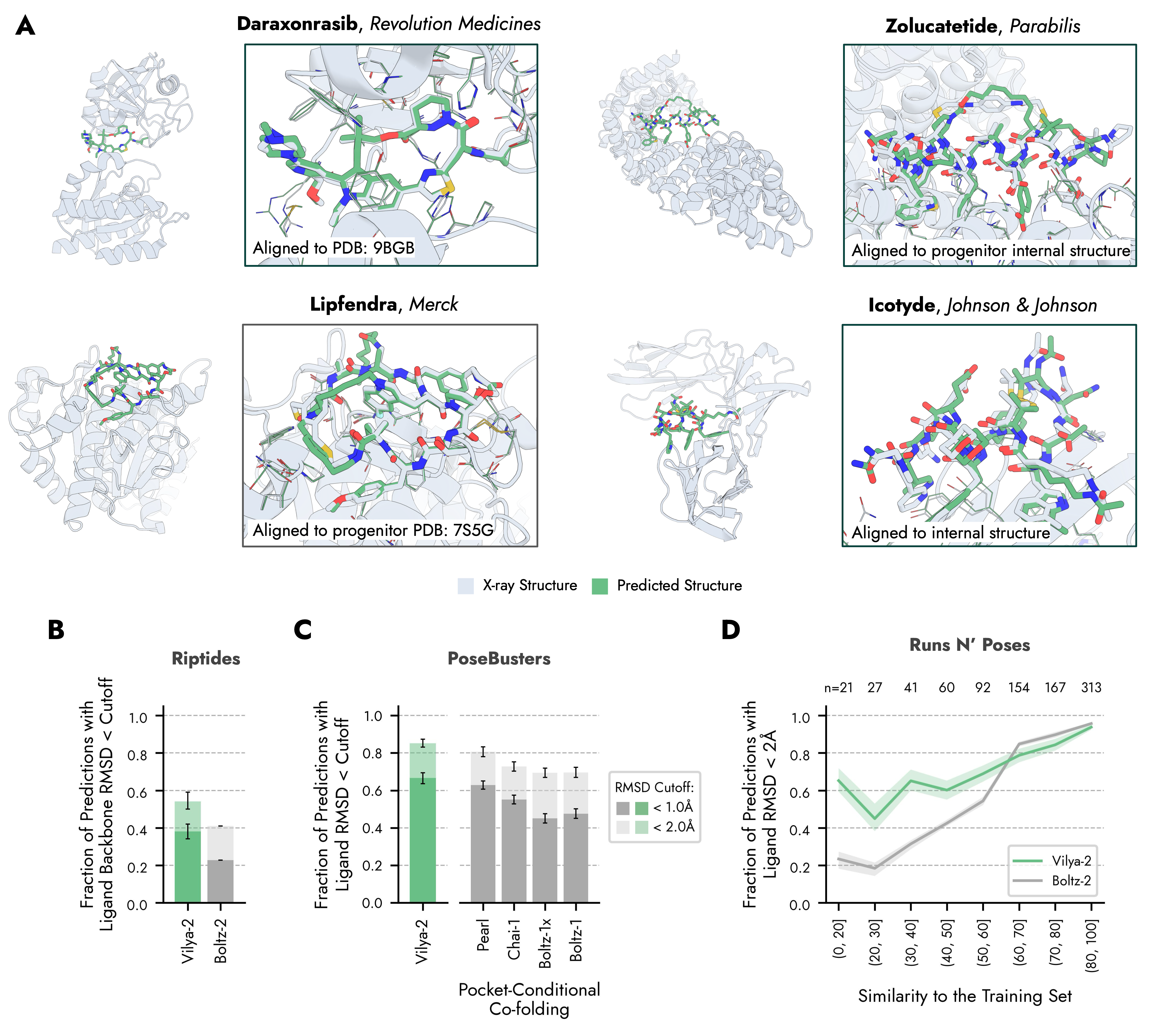}
  \caption{Accurate prediction of protein-ligand interactions with Vilya-2. A) Structural superpositions of Vilya-2 interaction predictions (green) with reference X-ray crystal structures (gray). Examples include two macrocyclic drug candidates (daraxonrasib and zolucatetide) and two approved macrocyclic drug (Lipfendra and Icotyde). Predictions for daraxonrasib and Icotyde are superimposed on their respective co-crystal structures (PDB: 9BGB and an internal X-ray crystallographic structure). Predictions for Lipfendra and zolucatetide are evaluated against the co-crystal structures of their respective progenitor molecules (PDB: 7S5G and an internal X-ray crystallographic structure). B) Performance on the Riptides benchmark. The fraction of predictions achieving a ligand backbone root-mean-square deviation (RMSD) $<$ 1.0 \AA~(dark bars) and $<$ 2.0 \AA~(light bars) is shown for Vilya-2 and the representative co-folding architecture, Boltz-2. C) Small molecule docking performance on the PoseBusters benchmark. Vilya-2 prediction success is compared against four co-folding baselines (Pearl, Chai-1, Boltz-1x, and Boltz-1 \cite{dobles2025pearl, wohlwend2025boltz, chai2024chai}). Baseline metrics are reproduced from \citet{dobles2025pearl}. In the pocket-conditional regime, co-folding methods are provided with the reference receptor crystal structure and specified binding pocket residues as inputs. D) Generalization capabilities evaluated on the Runs N' Poses benchmark. Prediction success rate (fraction of predictions with ligand RMSD $<$ 2.0 \AA) is plotted as a function of the interface maximum similarity to the training set for Vilya-2 (green) and Boltz-2 (gray). In panels B and D, Boltz-2 was additionally given the receptor conformation from the ground-truth co-crystal structure as an inference-time template. }
  \label{fig:fig1}
\end{figure}

Advances in protein structure prediction and design are transforming experimental discovery of
protein-based therapies. Laborious immunization- or display-based discovery methods are being replaced by \textit{de novo} design coupled with rapid, small-scale synthesis and testing of AI-designed mini-proteins and antibodies \cite{yang2026past}. These \textit{de novo} design pipelines involve the orchestration of several neural networks, which in turn generate candidate sequences and their structures, and then ultimately filter the generated molecules to handfuls worth producing in the laboratory \cite{butcher2025novo,cho2025boltzdesign1,team2025zero}. Fundamentally, the success of these pipelines has been driven by the predictive power of biomolecular structure prediction networks that arbitrate which \textit{de novo} designed molecules to actually test experimentally \cite{zambaldi2024novo, abramson2024accurate, bennett2023improving}. These tools rely both on accurate structure prediction of biomolecular complexes and model quality estimation to determine which AI-designed molecules may actually interact with the target protein in the wet lab.  

Existing biomolecular structure prediction networks have two limitations which make it challenging to apply them to designing and screening novel peptide candidates, especially peptides with non-canonical residues. First, they struggle to generalize to novel poses and pockets. Previous literature has shown that the performance of these tools depend heavily on the presence of similar protein-ligand interactions in their training sets, and/or the depth of evolutionary information available for homologous protein sequences \cite{vskrinjar2026evaluating, masters2025investigating}. Drug discovery campaigns often involve making novel interactions to a pocket not present in existing structural databases. Second, although modern networks are theoretically capable of representing arbitrary chemical inputs, in practice, they struggle to predict the structures of peptides with non-canonical chemistries. These failures range from errors in binding pose and orientation to more basic problems such as producing structures with incorrect chirality or implausible geometry \cite{zhang2025alphafold3}. These problems make \textit{de novo} design methods challenging to apply to peptide drug discovery: the majority of approved peptide drugs involve significant non-canonical modifications, and in some cases more esoteric chemical linkages that impose difficult-to-model topological constraints \cite{hickey2023beyond}.

In this paper, we introduce Vilya-2: a neural network designed for arbitrary structural modeling of both individual molecules and molecular interfaces. The driving philosophy behind Vilya-2 is two-fold. First, we posit that a split residue\slash atom representation fundamentally limits existing networks. This split representation prevents transfer learning between different molecular classes in the Protein Data Bank, and also complicates the integration of structural data beyond the PDB. Together, these limitations cause models to memorize known interactions rather than learn the fundamental principles of molecular organization \cite{masters2025investigating}. Second, we hypothesize that existing co-folding networks do not adequately leverage inference-time scaling - which has been critical for solving complex problems in language and vision. Despite employing recycling and a diffusion process, existing co-folding networks tend to produce a narrow range of structures despite using a diffusion process, which limits their ability to operate on unseen interactions. Vilya-2 is explicitly designed for diversity: producing and scoring multiple poses allows the model to explore alternate solutions to complex interactions.

Vilya-2 is a diffusion transformer whose input is the chemical graph of the system---its atoms and bonds and their elementary chemical and stereochemical features. Crucially, no residue-level tokens or molecule-type annotations are used: a peptide, a small molecule, and their protein target are all represented identically, as atoms in a single graph. The diffusion process runs through the entire architecture rather than being confined to a module atop a separate conditioning trunk. Instead of pursuing full co-folding of entire protein chains, we formulate the interaction between protein targets and ligands as a local, flexible structural-modeling task. While narrower in scope, this framing is in line with most drug discovery campaigns in which both the general structure of the target and the epitope are known.

On the Riptides benchmark---a novel benchmark modeling large, chemically-diverse peptide structures---Vilya-2 shows excellent performance and is able to predict the majority of structures accurately (backbone RMSD $<$ 2\AA), unlike existing co-folding networks. Vilya-2 also demonstrates state of the art performance at small molecule docking. Critically, this performance is able to generalize to poses dissimilar to those seen in training. Vilya-2 is able to predict structures of large macrocycles and miniproteins despite having no residue or evolutionary information, and generalizes to a variety of large, polymeric structures (Figure \ref{fig:fig1}). Vilya-2 can also be fine-tuned on experimental data to predict compound activity, which we show using measurements from an internal hit-to-lead optimization program. Overall, Vilya-2 represents a step-wise advance in generalization relative to existing biomolecular modeling tools and enables the modeling of a broad class of peptide therapeutics.

\section{Methods}

\subsection{The Vilya-2 Architecture}
The Vilya-2 architecture builds upon the unified all-atom transformer framework introduced in Vilya-1 \cite{vilya1}. A chemical graph with nodes representing heavy atoms and edges representing covalent bonds serves as the primary input to the network. The model embeds only fundamental chemical features derived from the graph (atomic number, hybridization state, formal charge, bond order, and conjugation\slash aromaticity state), explicitly avoiding residue-level tokenization, molecule-type annotations, or information from multiple sequence alignments. Additionally, node-level vector features are provided to inform correct local geometry and tetrahedral stereocenter configurations \cite{krishna2024generalized, anishchenko2025modeling}.

\begin{figure}[htbp]
  \centering
  \includegraphics[width=1.0\linewidth]{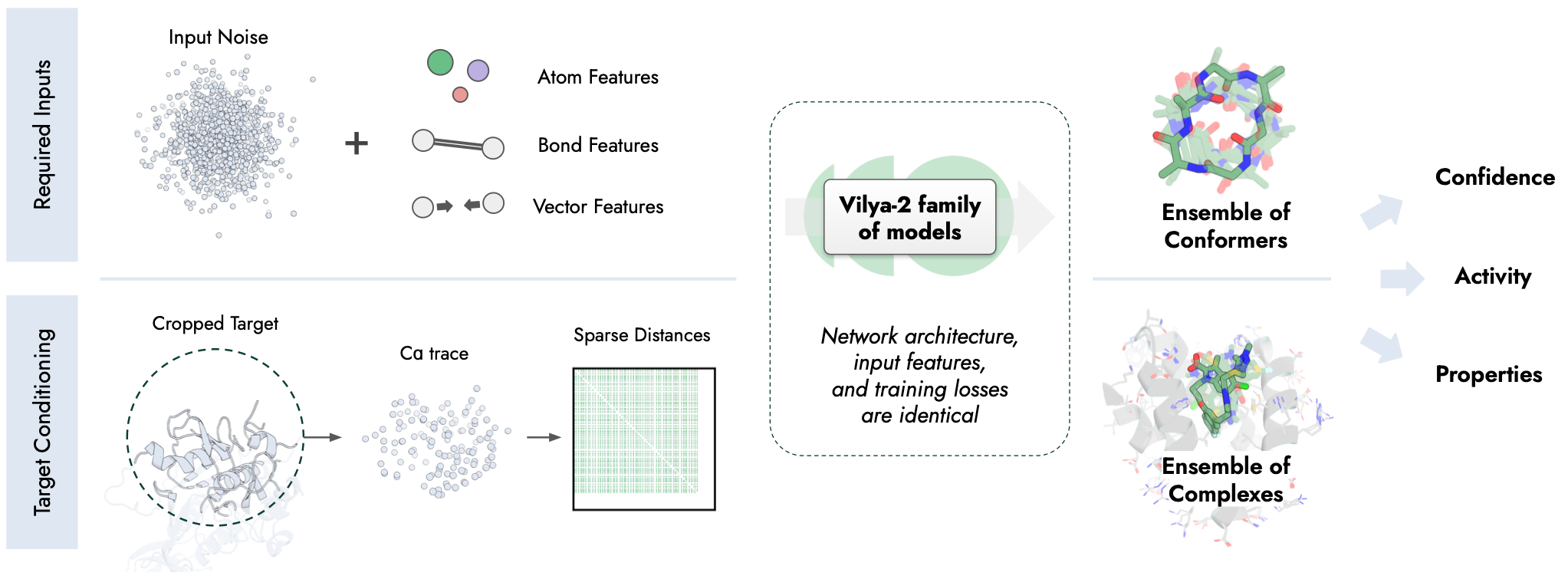}
  \caption{ An overview of Vilya-2. Structure-prediction networks form the core of Vilya-2: a conformer generator and a target-conditioned complex predictor, which generate ensembles of conformers or of protein–ligand complexes respectively. The two share the same architecture, input features, and training losses, and are trained first. Their architecture is then reused for specialized heads, which are fine-tuned to perform downstream predictions of confidence, activity, and molecular properties (right). }
  \label{fig:fig2}
\end{figure}

We modified the Vilya-1 architectural baseline to improve computational efficiency and scalability in Vilya-2. First, we removed all triangle attention layers as previous work observed better scaling relying solely on triangle multiplication for 2D updates \cite{ouyang2025triangle}. Second, inspired by recent literature scaling protein generative models, we increased the number of 1D track updates relative to the number of 2D track updates \cite{geffner2025proteina}. Finally, we moved dynamic index calculations that are required for vector feature updates outside of the core network. This enables full-graph compilation, accelerating overall computational throughput several-fold. An overview of the network is shown in Figure \ref{fig:fig2}.

\subsection{Training}
\paragraph{Conformer generation:} The training protocol for the conformer generator closely follows the methodology established for Vilya-1 \cite{vilya1}. The model is trained to reconstruct atomic coordinates from chemical inputs and noisy coordinates via a diffusion-based objective, augmented by two auxiliary losses: a pairwise distogram loss and a stereochemical loss to encourage correct tetrahedral chirality (Figure \ref{fig:fig2}, upper track). To train Vilya-2, the base training datasets utilized for Vilya-1 were expanded with additional peptide structures curated from the CPSea database \cite{yang2026cpsea}. 

\paragraph{Interface prediction:} The interface prediction model utilizes the same core training objective and chemical input features as the conformer generator, representing the entire molecular system (including the target protein, ligand, and cofactors) as a unified chemical graph. To the baseline set of atomic scalars, bond scalars, and vector-valued geometric features, we add one supplemental input to provide global structural context for the target. Specifically, a sparse distance matrix derived from non-ligand protein C$\alpha$ and nucleic acid C4' coordinates in the reference crystal structure is supplied as an additional 2D feature (Figure \ref{fig:fig2}, lower track). No additional molecule-type annotations are used to differentiate between target and ligand. This backbone-conditioned approach is grounded in the observation that side-chain flexibility is frequently the primary driver of binding dynamics in drug discovery \cite{clark2019inherent}.

During training, the input system is explicitly cropped to a 1024-atom interaction region centered on the ligand. This corresponds to approximately 120 protein residues, roughly the average size of a functional domain \cite{cheng2014ecod}. To prevent the network from trivially memorizing the location of the ligand within the binding site, the center of the crop is spatially jittered prior to cropping, following a distribution with a median radial displacement of 3.2 \AA. Every residue is then sorted via minimum atomic distance to the perturbed center of mass and iteratively added to the crop up to 1024 atoms. 

The training data was augmented beyond the conformer generation datasets by incorporating biomolecular complexes from the PDB and protein-peptide complexes from the CPSea database \cite{yang2026cpsea}. Rigorous parsing pipelines were implemented to ensure the accurate construction of complete chemical graphs directly from the deposited structural coordinates. To facilitate unbiased comparisons with existing co-folding methods, training was restricted to PDB entries released on or before 2021--09--30. The training losses remain identical to the conformer generation case. 

\paragraph{Confidence estimation:} After training the complex prediction model, we train a confidence estimation model to predict the local Distance Difference Test (lDDT) score \cite{mariani2013lddt}. To generate training data for this model, we sample a subset of the training dataset and generate 100 candidate poses per example. The confidence model is then trained using a cross-entropy loss to predict binned per-atom lDDT. Following Vilya-1, the Vilya-2 auxiliary heads share the same backbone architecture as the structure-prediction module. This design choice allows us to initialize the confidence model directly from the pre-trained structural checkpoint, thereby fully leveraging its learned representations. The confidence estimation model takes in no additional embeddings or inputs from the complex prediction model other than the predicted atomic coordinates of the system and features derived from the chemical graph.

\paragraph{Activity prediction:} Once Vilya-2 structure and confidence models are established, we train an activity head to predict target-specific experimental potencies of compounds from an internal hit-to-lead campaign. Unlike the confidence head, which takes a single conformer as input, the activity prediction head is supplied with an ensemble of conformers generated by the structure-prediction module. We train two variants of the activity predictor: one conditioned on ligand-only structures sampled by the Vilya-2 conformer generator, and the other on the co-complex ensembles generated by the Vilya-2 complex predictor. For each compound, we generate an ensemble of 1000 conformers (either ligand-only or target-bound complexes), initializing the activity head weights with those of the corresponding structure-prediction module.

We adapt the fine tuning protocol established in Vilya-1 by introducing three key modifications: (1) parameter-efficient fine-tuning via low-rank adaptors \cite{hu2022lora} to handle the larger pre-trained Vilya-2 models; (2) gated attention pooling \cite{ilse2018attention} to focus on the specific atoms driving molecular-level predictions in large atomic systems; and (3) conformation ensemble-level pooling to aggregate predictions across the generated structures and capture conformational variations. The model is trained to regress potency values obtained from an internal homogeneous time-resolved fluorescence (HTRF) assay \cite{degorce2009htrf}. Data was split chronologically to simulate the prospective performance of the model as the campaign progresses \cite{landrum2023simpd}.

\subsection{Evaluation Sets} \label{subsec23:evaluation_sets}
To evaluate our model’s performance at small-molecule docking, we rely on three recently developed benchmarks: PoseBusters \cite{buttenschoen2024posebusters}, Runs N’ Poses \cite{vskrinjar2026evaluating}, and PoseX \cite{jiang2025posex}. These benchmarks assess a network's performance at self- and cross-docking, physical validity of generated poses, and their generalization to novel protein-ligand systems.

We additionally introduce a novel benchmark, the Really Important Peptides (Riptides) benchmark - to test networks’ ability to model protein-peptide interfaces accurately. This benchmark is meant to establish a clear testing ground to assess the performance of structure prediction tools –be they large neural networks like Vilya-2 or classic physics-based methods– at modeling types of peptides that the peptide community develops as therapies. We constructed this benchmark with three goals in mind. First, the benchmark should contain a diversity of peptide topologies, both linear and cyclic peptides of myriad sizes, as the definition of a peptide-based therapy is broad (see Figure \ref{fig:fig1}A). Second, the benchmark should represent the chemical diversity that peptide chemists deploy during optimization by including examples with non-canonical amino acids. Third, the benchmark should undergo rigorous quality control to avoid evaluating methods on problematic or partially unresolved structures, which are prevalent in the PDB and may artificially degrade model performance \cite{dauter2014avoidable, deller2015models, wlodawer2018detect}.

Riptides includes both linear peptides and macrocycles. For the purposes of this benchmark, we select linear peptides from polypeptide chains that have between 6 and 18 residues. Macrocycles are chains annotated as either polypeptide or non-polymer, and have a cycle of at least 12 atoms. For the benchmark, we select macrocycles containing 1 to 20 residues. This definition of a macrocycle is broad; it includes myriad topologies and is not constrained to simple head-to-tail cyclization. We select structures from the chains in the PDB that satisfy the following filters:

\begin{itemize}
    \item The release date of the entry must be past 2023-06-01. This is done in order to fairly compare against Boltz-2 \cite{passaro2025boltz}.
    \item The resolution of the entry must be $<$ 3\AA.
    \item There must be at least 20 unique atom pairs with a distance $<$ 5\AA~between the ligand and its closest protein receptor chain. Receptor chains must have at least 50 resolved residues.
\end{itemize}

\begin{figure}[htbp]
  \centering
  \includegraphics[width=1.0\linewidth]{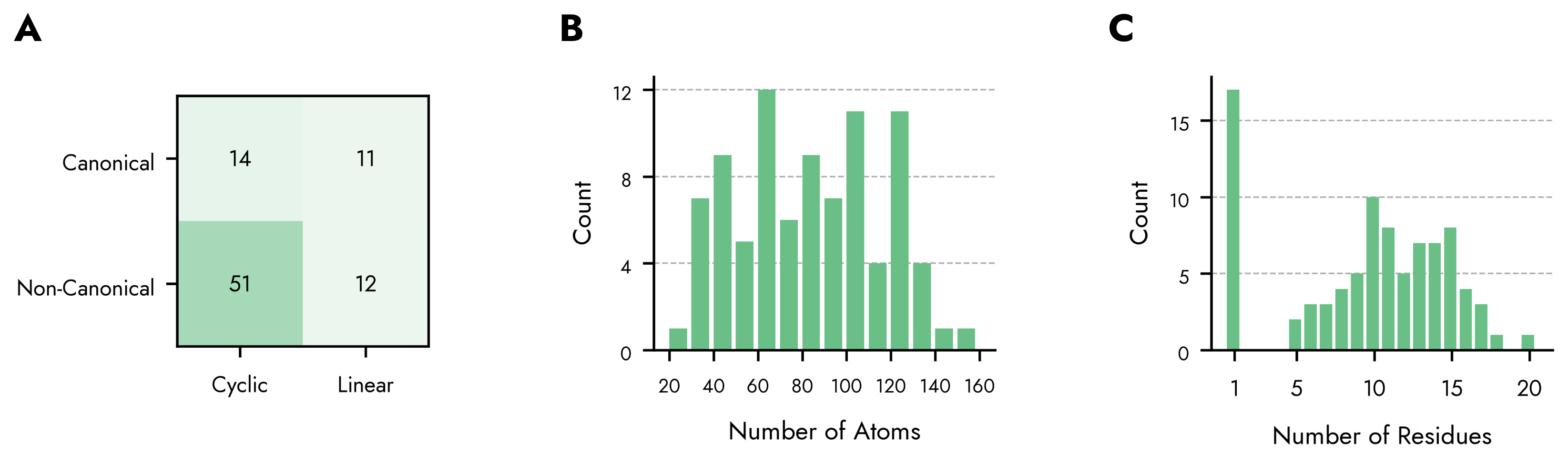}
  \caption{ Composition of the Riptides benchmark. A) Breakdown of the 88 entries in the benchmark by topology (cyclic vs linear) and chemical composition (canonical vs non-canonical) with cell shading proportional to count; non-canonical cyclic peptides dominate the set (51 of 88). B) Distribution of ligand size, in number of atoms. C) Distribution of ligand length, in number of residues; macrocycles annotated as non-polymers in the PDB are counted as a single residue (spike at 1). }
  \label{fig:fig3}
\end{figure}

We then deduplicate the resulting entries by primary protein receptor and ligand chemistry, and manually drop entries with unresolved residues and entries where the ligand is making inadequate contact with the primary protein receptor. Our final benchmark contains 88 examples. Basic statistics about this set are in Figure \ref{fig:fig3}. Each entry is then prepared using Schr{\"o}dinger's Protein Preparation Workflow with maximum atom count for epikx increased to 500 with all other default settings \cite{madhavi2013protein, schrodinger2026protprep}. Structures were stripped of all waters and all chains far from the binding site, including cofactors and metal ions. Some charged states in the ligands were manually corrected. N and C termini on both the receptor and peptide ligands were processed as follows:

\begin{itemize}
    \item Termini representing the true N- and C- terminal ends were converted to ammonium and carboxylate ions, respectively, to reflect their charged state at physiological pH.
    \item N- and C- termini representing peptide bonds adjacent to unresolved atoms in the crystal structure were capped with acetyl or N-methyl groups, respectively.
\end{itemize}

\subsection{Evaluation Methodology}

Unless otherwise specified, all interface predictions are done by sampling 100 poses via a diffusion process. Each pose is independently ranked by the confidence estimation model. Because the model outputs a categorical distribution over discrete pLDDT bins, we first calculate the expected value of the pLDDT for each individual atom. These per-atom expectations are then averaged across the atoms of interest. For the evaluations presented in this paper, the final pLDDT scores are averaged exclusively over the ligand atoms (plddt-ligand).

The top-ranked prediction is evaluated against the experimental crystal structure. In Figure \ref{fig:fig4}C, we generate 10k samples to better map the conformational landscape of the four complexes of interest; predictions with the highest plddt-ligand are visualized in Figure \ref{fig:fig1}A. In Figure \ref{fig:fig4}A, we also plot ``Oracle'' ranking, which is using the lowest RMSD pose generated out of $n$ samples. In practice, this sort of ranking is unrealistic as the ground-truth structure is not known. However, it serves to show the gaps between selecting poses uniformly at random (equivalent to sampling a single pose), our scoring model, and the theoretical upper bound on scoring. 

Confidence intervals for all structure prediction results are obtained by bootstrapping. In particular, for Figure \ref{fig:fig1}B-D, we generate and score a total of 1000 poses for each input system. We then select 100 poses uniformly at random and take the highest plddt-ligand prediction for each system. We compute the success rate as the fraction of highest-ranked poses under some ligand RMSD threshold, and then repeat this process 1000 times. We then plot the mean success rate and 95\% confidence intervals.

When running Boltz-2 as a baseline, we use Boltz's reported ipTM as the scoring metric. In practice, however, we find no significant difference between different metrics (\verb|confidence_score|, \verb|iptm|, \verb|ptm|, or any variant of subselecting the predicted \verb|pae| matrix to only protein-ligand interactions). In order to fairly compare against Boltz-2, we provide the receptor’s experimental backbone coordinates as templates in addition to the sequence, which allows us to more appropriately measure the performance of Boltz-2 specifically at interface modeling. We omit other co-folding models from our internal evaluations with the observation from recent literature that the performance of these models is generally correlated \cite{vskrinjar2026evaluating}.

Structural alignment between a predicted pose and the experimentally resolved structure is done based on the N--C$\alpha$--C atoms of the protein receptor, and only performed on residues with at least one atom within 10\AA~of any atom of the ligand. Up to 128 possible symmetric permutations of the ligand are enumerated during RMSD calculations. For linear peptides and macrocycles, we evaluate both the full atom RMSD and the backbone RMSD. We define the set of backbone atoms $S$ as follows:

\begin{itemize}
    \item Initialize an empty set of atoms $S$.
    \item For macrocycles, add to $S$ all atoms in the largest cycle (in the minimum cycle basis, i.e. smallest set of smallest rings).
    \item For linear peptides, add to $S$ all atoms matching the backbone fragment \verb|NCC=O|.
    \item For every other cycle $C$ in the minimum cycle basis, add every atom in $C$ to $S$ if $C \cap S \neq \emptyset$.
    \item For every atom $x \in S$, add every neighbor of $x$ to $S$.
\end{itemize}

Backbone RMSD is measured on every atom in $S$. Illustrations of the backbone set of atoms are in Figure \ref{fig:fig_s3}. This expanded definition of backbone requires correct placement of the core scaffold of the ligand while excluding highly flexible side-chains and tails that may not always adopt a unique conformation.

\section{Results}
\subsection{Interface Prediction}

\paragraph{Vilya-2 is state-of-the-art at peptide interface prediction.} We first assess Vilya-2's ability to recover the bound conformation of a peptide at the protein-target interface. This is the task most directly relevant to peptide drug discovery, in which a structure-prediction network serves as the final oracle in a de novo design pipeline.

\begin{figure}[htbp]
  \centering
  \includegraphics[width=1.0\linewidth]{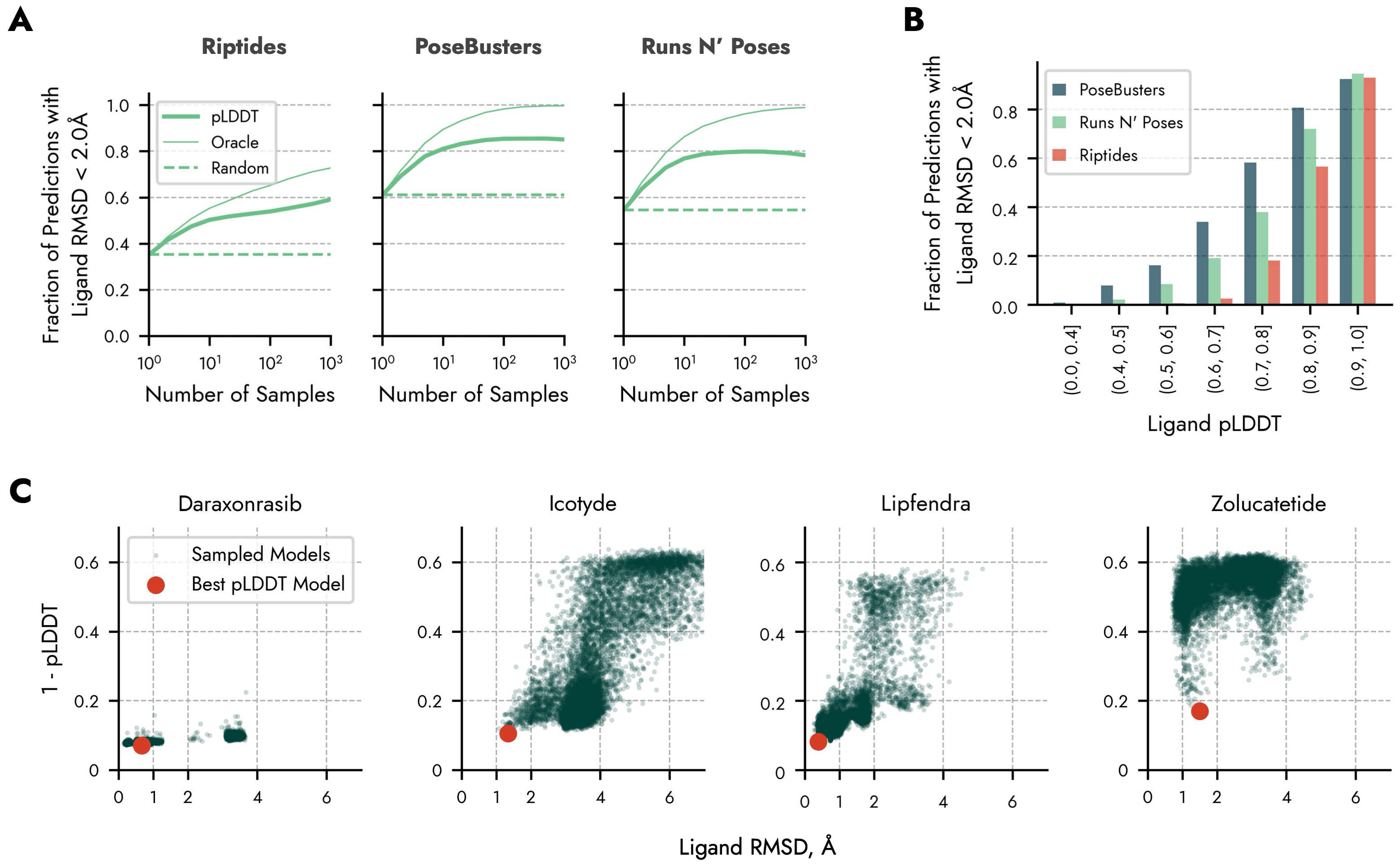}
  \caption{ Interplay between efficient sampling and accurate scoring drives Vilya-2 predictive accuracy. A) Increased sampling combined with pLDDT-based ranking yields higher prediction success. The fraction of targets with a successfully predicted bound ligand conformation (RMSD $<$ 2\AA) is plotted as a function of the number of generated samples. Success rates are evaluated using top-ranked selection based on plddt-ligand (thick line), ideal oracle selection (thin line), and random selection (dashed line) across the Riptides, PoseBusters, and Runs N’ Poses datasets. B) Empirical calibration of the Vilya-2 confidence metric. The prediction success rate (fraction of models with ligand RMSD $<$ 2\AA) is plotted across discrete bins of plddt-ligand scores for the three benchmark datasets, demonstrating a strong correlation between the predicted confidence score and the probability of a structurally accurate prediction. C) Scoring funnels for the targets shown in Figure \ref{fig:fig1}A. Across all four targets, the sampled structural ensembles (dark green circles) funnel toward the native conformation; consequently, ranking by the plddt-ligand metric successfully identifies a low-RMSD state as the top model (red circle). Because Lipfendra and zolucatetide are evaluated against co-crystal structures of a progenitor molecule, their RMSD is calculated over atoms belonging to the maximum common substructure between the two molecules. All-atom RMSD is used for daraxonrasib and Icotyde. }
  \label{fig:fig4}
\end{figure}

Evaluated on the Riptides Benchmark described in Section \ref{subsec23:evaluation_sets}, Vilya-2 recovers the bound peptide backbone in 54.1\% of cases at a stringent $<$ 2\AA~RMSD and 38.0\% at $<$ 1\AA~(Figure \ref{fig:fig1}B and Table \ref{tab:tab1})---thresholds considerably more demanding than those typically used to report peptide-interface accuracy elsewhere (for example, DockQ $>$ 0.23 in \cite{qiao2026neuralplexer3}). The representative co-folding model Boltz-2 reaches only 40.9\% and 22.7\% at the same thresholds, despite being conditioned on the receptor crystal structure as a template and having an extra year and a half of training data. This performance gap widens further with increased sampling: at 1000 samples, Vilya-2's accuracy rises to 59.1\% at $<$ 2\AA~and 40.9\% at $<$ 1\AA~(Figure \ref{fig:fig4}A).

We also benchmarked Vilya-2 against a representative physics-based approach using Schrödinger's docking tools, applying MacroDock \cite{robson2026accurate} to macrocycles and Glide \cite{friesner2004glide} to linear peptides. For each target we generated 100 poses, ranked them by docking score, and selected the top-ranked model. Schrödinger docking recovers the bound peptide backbone in only 22.0\% of cases at $<$ 2\AA~and 15.9\% at $<$ 1\AA---more than twofold lower than Vilya-2 at both thresholds (Figure \ref{fig:fig_s4}).

\begin{table}[ht]
\centering
\caption{ Vilya-2 performance across the Riptides, PoseBusters, Runs N’ Poses, and PoseX benchmarks. Values are success rates (the percentage of predictions falling below the indicated RMSD cutoff); cutoffs are computed over backbone atoms for the Riptides Benchmark and over all heavy atoms for the other three. Bracketed values are 95\% confidence intervals from bootstrapping (see Methods). }
\label{tab:tab1}
\begin{tabular}{lcccc}
\toprule
 & RMSD cutoff & PB-Valid = False & PB-Valid = True & PB-Valid prefilter \\
\midrule
\multirow{2}{*}{Riptides} & 1\AA{} & 38.0 $\pm$ 4.0 & 35.7 $\pm$ 5.1 & 39.0 $\pm$ 4.1 \\
 & 2\AA{} & 54.1 $\pm$ 4.6 & 48.7 $\pm$ 4.6 & 53.7 $\pm$ 4.9 \\
\midrule
\multirow{2}{*}{PoseBusters} & 1\AA{} & 66.7 $\pm$ 3.1 & 65.6 $\pm$ 2.9 & 66.8 $\pm$ 3.3 \\
 & 2\AA{} & 85.2 $\pm$ 2.1 & 83.5 $\pm$ 2.5 & 85.3 $\pm$ 2.0 \\
\midrule
\multirow{2}{*}{Runs N' Poses} & 1\AA{} & 61.0 $\pm$ 1.7 & 58.3 $\pm$ 1.8 & 61.5 $\pm$ 1.8 \\
 & 2\AA{} & 79.5 $\pm$ 1.2 & 75.6 $\pm$ 1.4 & 80.3 $\pm$ 1.2 \\
\midrule
\multirow{2}{*}{PoseX-SD} & 1\AA{} & 57.5 $\pm$ 1.8 & 55.6 $\pm$ 2.0 & 57.5 $\pm$ 1.9 \\
 & 2\AA{} & 76.0 $\pm$ 1.6 & 72.4 $\pm$ 1.6 & 75.3 $\pm$ 1.5 \\
\midrule
\multirow{2}{*}{PoseX-CD} & 1\AA{} & 47.8 $\pm$ 1.4 & 45.9 $\pm$ 1.4 & 48.5 $\pm$ 1.3 \\
 & 2\AA{} & 71.1 $\pm$ 1.1 & 66.5 $\pm$ 1.3 & 70.8 $\pm$ 1.2 \\
\bottomrule
\end{tabular}
\end{table}

Beyond aggregate benchmarks, Vilya-2 accurately models individual therapeutic macrocycles of direct clinical interest (Figure \ref{fig:fig1}A). Vilya-2 recapitulates the bound conformations of two macrocyclic drug candidates---daraxonrasib and zolucatetide---and the approved macrocyclic drugs Lipfendra and Icotyde (Figure \ref{fig:fig1}A). It is worth emphasizing that in all of these cases, none of the experimental structures we compare Vilya-2's predictions against were present in the training data on which the model was trained. Vilya-2's ability to predict these structures emphasizes its ability to handle arbitrarily complex chemistries and topologies. The helicon zolucatetide adopts an $\alpha$-helix that is enforced by three separate side-chain to side-chain crosslinks: an $i$,$i+8$ linkage via a long hydrocarbon chain an $i$,$i+4$ linkage via a short hydrocarbon chain, and an $i$,$i+5$ linkage via an amide bond. While there is no public co-structure of zolucatetide bound to $\beta$-catenin to corroborate our predicted complex, we observe near identical interactions between an internally solved costructure of a progenitor of helicon and the predicted complex from Vilya-2. The progenitor compound in this instance is similar in chemical structure to those previously reported by Parabilis yet binds at the TCF4:$\beta$-catenin interface \cite{li2022novo}. The disulfide lariat peptide drug Icotyde \cite{fourie2024jnj77242113} lacks the complex linkage scheme present in zolucatetide yet Vilya-2 predicts its lariat tail to adopt an $\alpha$-helix in complex with IL23R. Undoubtedly the $\alpha$-disubstituted oxane unnatural amino acid acts to stabilize this helical conformation\cite{marshall1990factors}. Our internally resolved co-structure of Icotyde bound to IL23R near perfectly matches the Vilya-2 prediction with only very subtle differences in side chain chi angles across the peptide. Our daraxonrasib prediction demonstrates that Vilya-2 can handle topologies that are minimally peptidic; daraxonrasib contains only 2 amide bonds, one involving a hydrazine, whereas the rest of the molecule is fashioned from moieties more common in small molecules. Vilya-2's prediction of this molecule is nearly identical to Revolution Medicines' reported co-structure available in the PDB \cite{cregg2025discovery}. The Vilya-2 prediction of Lipfendra is remarkable as this peptide contains no defined secondary structure and is instead held in the binding competent confirmation via very elaborate crosslinking strategies \cite{alleyne2020series, tucker2021series}. We observe that the predicted complex perfectly recapitulates the interactions that a progenitor compound makes in complex with PCSK9. 

\paragraph{Vilya-2 generalizes to structures not seen at training.} The practical utility of a structure prediction model depends on its ability to generalize to novel pockets and binding modes. The performances of co-folding networks have been shown to drop substantially on targets dissimilar to their training data, which limits their applicability in novel drug-discovery campaigns.

We evaluate Vilya-2 on the Runs N’ Poses benchmark \cite{vskrinjar2026evaluating}, which explicitly measures generalization performance of machine learning methods trained on the PDB. We select the PDB entries with a release date after 2023--06--01 for fair comparison with Boltz-2, and compute similarities with respect to that cut-off date. Note that Vilya-2 is still trained on the earlier cutoff date of 2021--09--30. We sample poses separately for every proper ligand in the benchmark and include all co-factors around that ligand within a 1024 atom crop. We additionally run Boltz-2 with the inputs and script provided by the benchmark, with modifications for sampling 100 poses and adding crystal-templating of the receptor structures as input. We observe, however, that Boltz-2's performance is largely independent of this template conditioning, showing minimal change in success rate when compared to default settings (Figure \ref{fig:fig_s1}).

As shown in Figure \ref{fig:fig1}D, Vilya-2 shows moderate decline in performance with respect to training similarity, maintaining strong performance ($>$ 60\% success rate at ligand docking) for the majority of similarity ranges on the benchmark, including the lowest. This is in contrast to Boltz-2, which drops to 20\% at the lowest evaluated similarity range. This performance is indicative of strong generalization: Vilya-2 is able to predict and rank small molecule poses unlike those it has seen in training. 

\paragraph{Vilya-2 produces physically-valid poses.} Many existing machine learning models have another failure mode: producing physically implausible structures that violate basic chemical and geometric rules \cite{buttenschoen2024posebusters}. Applying the PoseBusters filtering checks can drop success rates of co-folding methods by up to 10\% on certain benchmarks, and success rates for other docking models can drop by up to 50\% \cite{jiang2025posex}. Full results for our method with PoseBusters filtering are in Table \ref{tab:tab1}. We find that the top-ranked prediction satisfies PoseBusters filters the majority of the time---Vilya-2 only drops by $\sim 4 \%$ points when applying these filters.

We make the additional observation that, because the PoseBusters filters depend solely on the chemical graphs of the input molecules and not on the true conformation of the bound structure, we can apply PoseBusters filters before ranking with our confidence model. This filtering scheme, which we denote as ``PB-valid prefilter'', allows us to discard poses that seem physically implausible and guarantee that the top-ranked output structure passes these checks. By doing so, we recover the same success rates as before---and even improve in success rate in some cases---while by definition satisfying 100\% of the PoseBusters checks. This pre-filtering succeeds due to the synergy between Vilya-2's high conformational diversity and baseline quality (most poses already satisfy the PoseBusters criteria), ensuring that every 100-sample pool contains multiple valid candidates for subsequent ranking.

\paragraph{Vilya-2 improves with sampling.} Success in modern vision and language models has been in-part driven by stochastic sampling and inference-time scaling, with the theory that harder problems require more computation and exploration to solve \cite{balachandran2025inference}. Scaling inference in co-folding methods requires re-running both the trunk and the diffusion module. While this has been shown to improve predictions for some interface types---notably antibody–antigen complexes (\cite{abramson2024accurate, isodde2026})---comparable gains have not, to our knowledge, been demonstrated for small-molecule or peptide interfaces. In contrast, Vilya-2 scales inference through the diffusion process alone. Figure \ref{fig:fig4}A  shows how Vilya-2 scales with diffusion samples: on small molecules success rates saturate at 100 samples, but for the more challenging peptidic interfaces, Vilya-2 continues to improve at 1000 samples.

The interplay between Vilya-2's sampling capability and its confidence network’s scoring accuracy enables us to mimic the ab initio-generated landscapes---colloquially referred to as forward folding funnels---commonly plotted by older physics-based methods \cite{huang2016coming}. Folding funnels for the peptides in Figure \ref{fig:fig1}A are shown in Figure \ref{fig:fig4}C. These results demonstrate the importance of both diverse sampling and accurate scoring. Vilya-2 samples a range of binding modes that are both near and far away from the near-native structure, and consistently identifies the near-native structure with confidence estimation. Sampling results for Boltz-2 on the Riptides benchmark are shown in Figure \ref{fig:fig_s2}. Co-folding methods like Boltz-2 suffer from two failure modes: first, that their sampled population is not diverse, and therefore their theoretical performance ceiling is limited. Second, and perhaps more alarmingly, their confidence estimation of ligand-protein interfaces is essentially no better than random. This phenomenon has also been observed for small molecule protein complexes in previous work \cite{dobles2025pearl}.

\paragraph{Vilya-2 is well-calibrated at scoring.} An important prerequisite for any generative structural modeling framework is a well-calibrated confidence metric. Ranking power alone is not enough: a scoring model may reliably order one structure above another yet give no absolute signal of whether a sampled population contains a near-native pose. Calibration provides this missing signal: a well-calibrated confidence score reflects how likely a prediction is to be near-native, indicating whether a good structure has already been found or more sampling is needed.

As demonstrated in Figure \ref{fig:fig4}B, the plddt-ligand score in Vilya-2 exhibits strong empirical calibration, showing that it can serve as a reliable proxy for indicating prediction success. For instance, across the three benchmark datasets, finding a pose with plddt-ligand $>$ 0.8 translates to a $>$ 60\% chance that the sampled ligand conformation falls under the full-atom RMSD of 2\AA. In contrast, a plddt-ligand score less than 0.6 is a strong indication that more sampling is required to find the correct binding mode. Calibration makes the model practically useful for novel ligands: it gives an absolute signal of when predictions are trustworthy.

\paragraph{Vilya-2 generalizes to cross-docking.} Vilya-2 is conditioned on C$\alpha$-C$\alpha$ pairwise distances of the receptor structure with the philosophy that for most drug discovery campaigns, the structure of the epitope is already loosely known. This input conditioning provides sparse, noisy distances about receptor fold to circumvent the need for learning residue-residue contacts from evolutionary information, but provides no information about side-chain packing. In practice, however, the exact bound conformation of the receptor is not known, raising an important question: to what degree can Vilya-2 actually cross-dock ligands into non-native receptor structures? 

Results on the PoseX benchmark, a benchmark designed to test the ability of existing machine learning methods to dock small molecules into non-native conformations of the receptor, are shown in Table 1 \cite{jiang2025posex}. The results demonstrate only a minor ($\sim 5 \%$) drop in success rate from the self-docking to the cross-docking scenarios, indicating that the backbone prior we use for the target leaves enough freedom to accurately model the majority of ligand-induced conformational changes. On both self and cross-docking, Vilya-2 outperforms all existing co-folding models (compare Table \ref{tab:tab1} results to Table S4 in \citet{jiang2025posex}). On cross-docking specifically, Vilya-2 succeeds on 71.1\% of targets (ligand RMSD $<$ 2\AA), falling only slightly to 66.5\% when poses must also pass physical-validity checks (PB-Valid). By comparison, AlphaFold 3 achieves 68.8\% and 53.8\% under the same two criteria, a much larger drop when physical validity is enforced.

\paragraph{Vilya-2 learns to predict potency in a hit-to-lead campaign.} Beyond structure prediction, Vilya-2 can serve as a foundation model: its pretrained representations can be efficiently fine-tuned for downstream tasks. We demonstrate this by adapting Vilya-2 to predict compound potency in a real-world hit-to-lead optimization campaign, using measured potency data from an internal program.

\begin{figure}[htbp]
  \centering
  \includegraphics[width=0.6\linewidth]{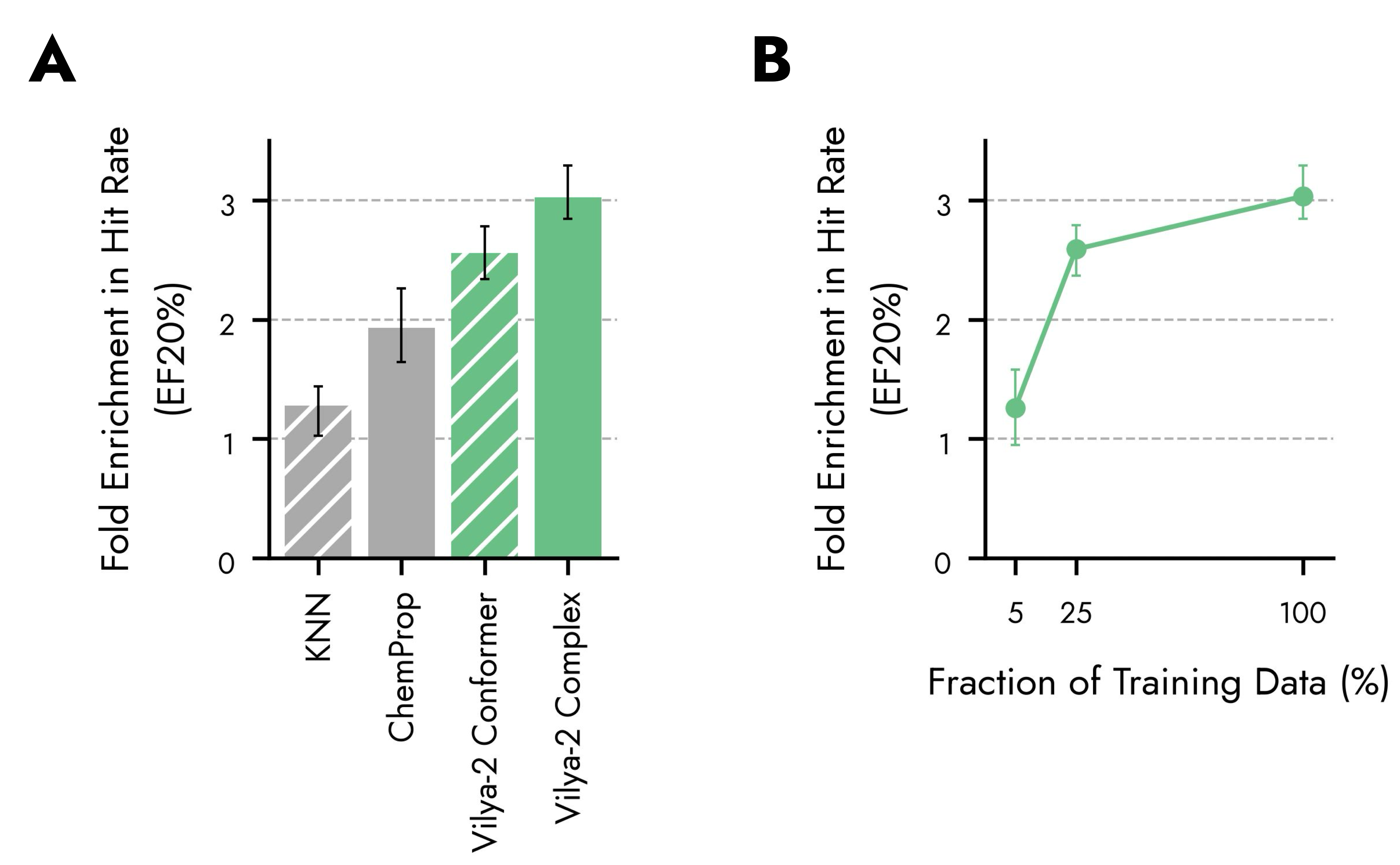}
  \caption{Activity Prediction with Vilya-2 on an internal hit-to-lead optimization campaign. A) Fold enrichment in hit rate among the top 20\% of ranked compounds (EF20\%) for two ligand-only baselines---K-nearest neighbors (KNN) on ECFP fingerprints and ChemProp---and the two Vilya-2 variants. The conformer-based model (light green hatched), fine-tuned on ligand conformers alone, and the complex-based model (light green), fine-tuned on generated ligand–target complexes, both outperform the baselines, with the complex-based model performing best. B) Effect of training-set size on the complex-based model: EF20\% as a function of the fraction of campaign data used for fine-tuning (5\%, 25\%, 100\%). Performance rises steeply and largely saturates by 25\% of the data. Error bars show 95\% confidence intervals. For reference, the maximum attainable EF20\% on this dataset is 4.875.}
  \label{fig:fig5}
\end{figure}

In a retrospective evaluation on this campaign, fine-tuning the complex-based Vilya-2 yields a $3\times$ improvement in enrichment for active compounds (Figure \ref{fig:fig5}A). The conformer-based variant (discussed in next section), which sees only the ligand, reaches a lower $2.5\times$ enrichment, suggesting that supplying the target context benefits learning. Both Vilya-2 variants base their predictions on 3D conformer ensembles: ligand conformations for the conformer-based model, and ligand–target interaction regions for the complex-based model. By contrast, the two baselines---K-nearest neighbors (KNN) on extended-connectivity fingerprints (ECFP) and ChemProp \citep{heid2023chemprop}---which predict from the 2D chemical structure of the ligand alone, show considerably lower enrichment ($1.3\times$ and $1.9\times$, respectively). Vilya-2 is also data-efficient: fine-tuned on fewer than a thousand early-campaign measurements, it approaches the performance of training on all available data (Figure \ref{fig:fig5}B).

\subsection{Improvements in Conformer Generation}

Having established Vilya-2's accuracy on biomolecular interfaces, the remainder of this work turns to conformer generation---sampling 3D structures of a molecule in isolation. This task serves two purposes. First, accurate conformer generation is foundational to downstream reasoning about molecules: the low-energy structures a molecule can adopt govern its binding and developability, so a faithful conformer model is the foundation for property prediction. Second, because it operates on single molecules rather than full complexes, the conformer generator is substantially faster to train and evaluate than the complex predictor, making it an efficient testbed for architectural exploration---the design improvements we identify here directly informed the interface model.

\begin{figure}[htbp]
  \centering
  \includegraphics[width=1.0\linewidth]{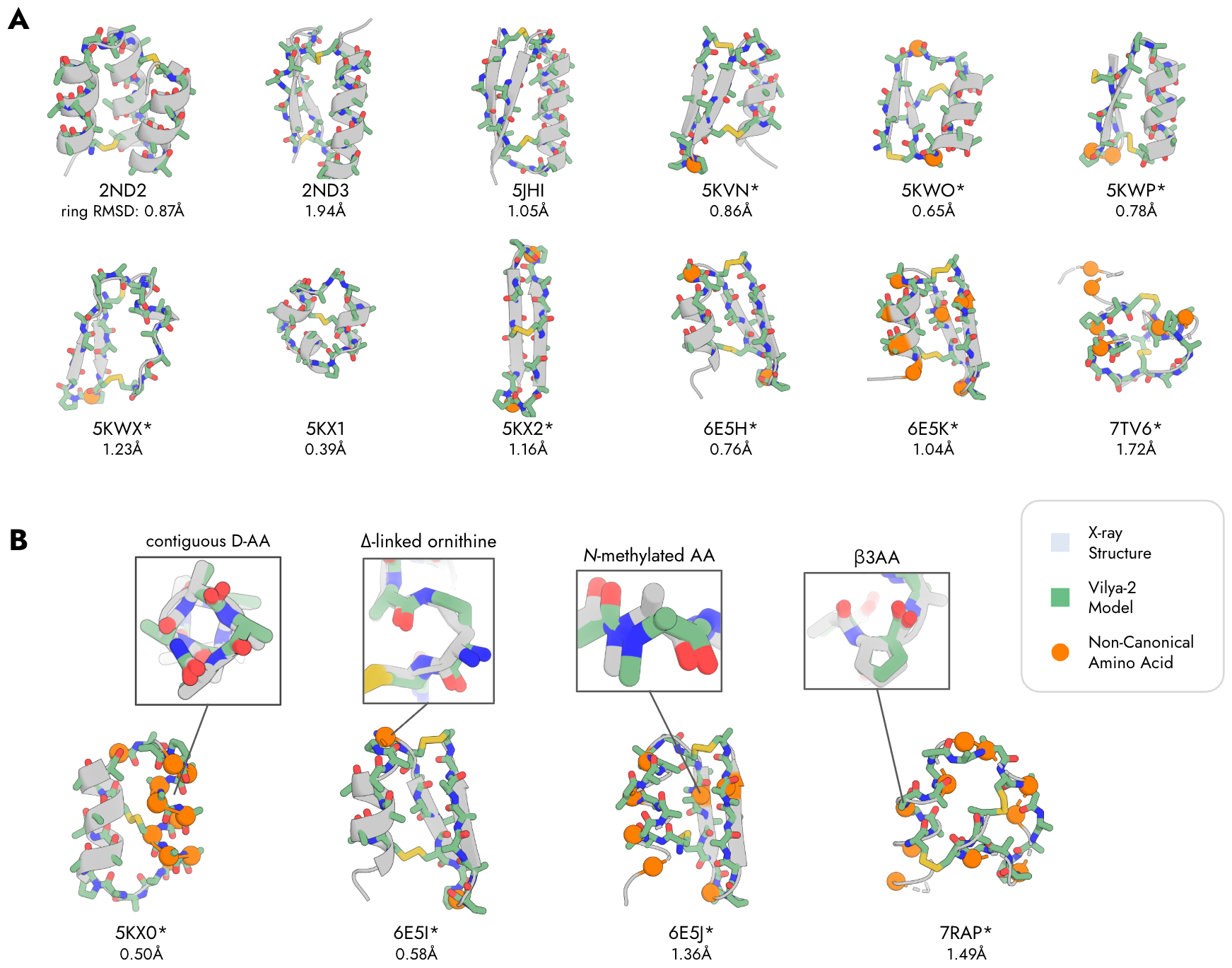}
  \caption{Vilya-2 accurately recapitulates structures of disulfide-stapled miniproteins. A,B) Overlays of experimental NMR structures (gray cartoon) with the best-sampled Vilya-2 conformation (green sticks; n = 4,000 generated samples). For clarity, Vilya-2 models are displayed using ring scaffold heavy atoms only, with side chains omitted. Asterisks (*) denote structures containing non-canonical amino acids, including D-amino acids (corresponding C$\alpha$ atoms are shown in orange spheres). Experimental NMR structures are sourced from three studies \citep{bhardwaj2016accurate, cabalteja2019heterogeneous, cabalteja2022heterogeneous}, which together comprise 21 disulfide-stapled miniproteins; panels A and B show representative examples, and ring RMSD values for the complete set are reported in Table \ref{tab:tabs1}. B) Additional examples from the same three studies, chosen to illustrate some of the non-canonical chemistries the network can model.}
  \label{fig:fig6}
\end{figure}

\begin{figure}[htbp]
  \centering
  \includegraphics[width=1.0\linewidth]{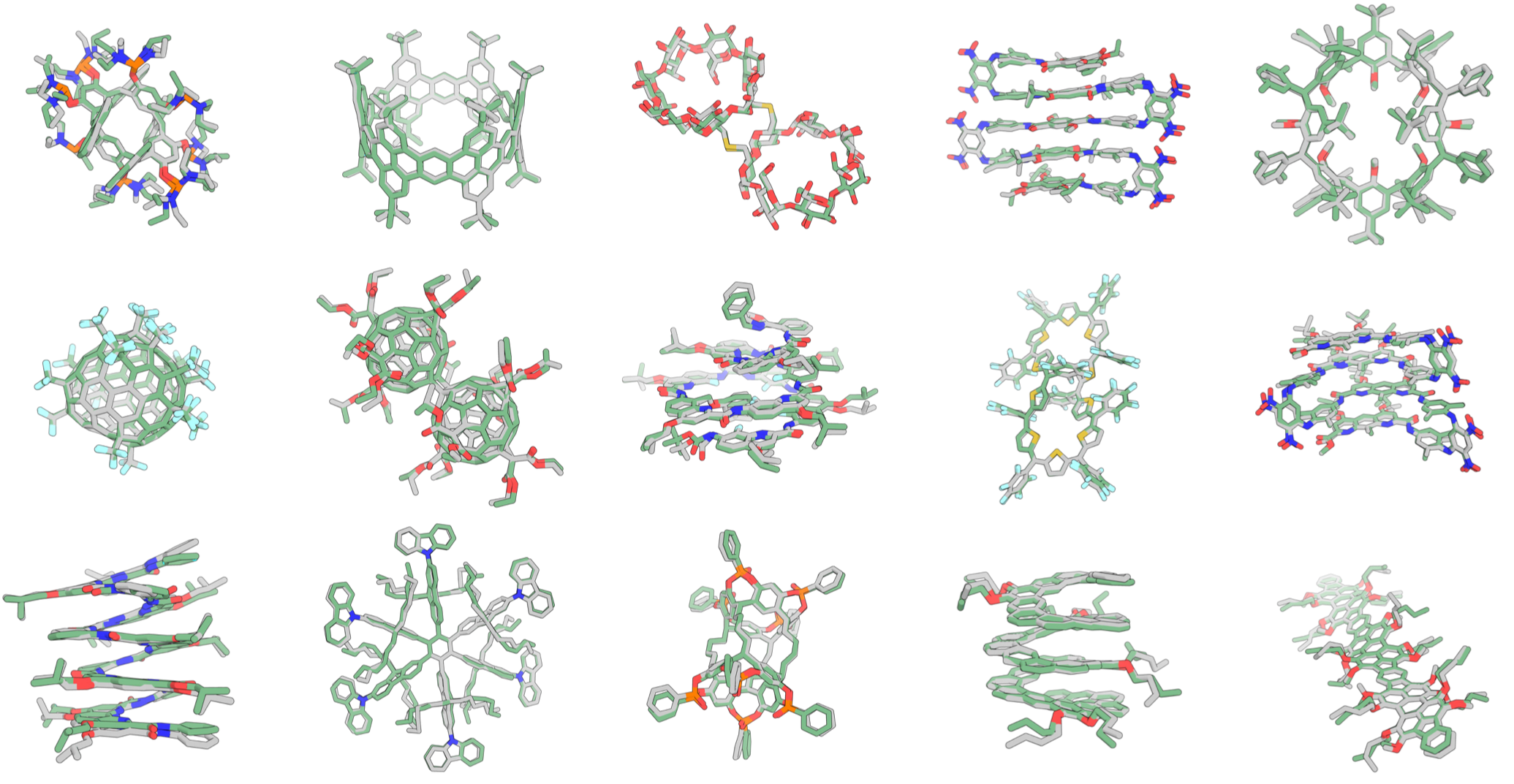}
  \caption{Vilya-2 accurately recapitulates structures of large organic molecules. Overlays of X-ray crystal structures from the Cambridge Structural Database (gray) with Vilya-2 models (green). All depicted molecules contain $>128$ heavy atoms. For each molecule, the model with the lowest all-atom RMSD to the crystal structure from an ensemble of 1,200 generated samples is shown. All displayed Vilya-2 models achieve an all-atom RMSD $<$ 1\AA.}
  \label{fig:fig7}
\end{figure}

\paragraph{Vilya-2 generalizes across molecular sizes and chemistries.} To evaluate the generalization capabilities of the Vilya-2 conformer generator, we tested the network on a diverse set of disulfide-stapled miniproteins (Figure \ref{fig:fig6}). During conformer generation training, Vilya-2 was limited to molecules containing less than 128 heavy atoms (mean = 31 heavy atoms). Despite this, the network accurately recapitulates the structures of miniproteins two to three times larger than any molecule present in its training set. Consistent with the performance of Vilya-1, Vilya-2 predicts structures with equivalent accuracy for both fully canonical peptides and highly modified sequences containing non-canonical residues. Furthermore, this size generalization extends beyond peptide-like scaffolds to diverse chemical classes, as demonstrated by the accurate structural prediction of large synthetic organic molecules (Figure \ref{fig:fig7}).

\begin{figure}[htbp]
  \centering
  \includegraphics[width=0.7\linewidth]{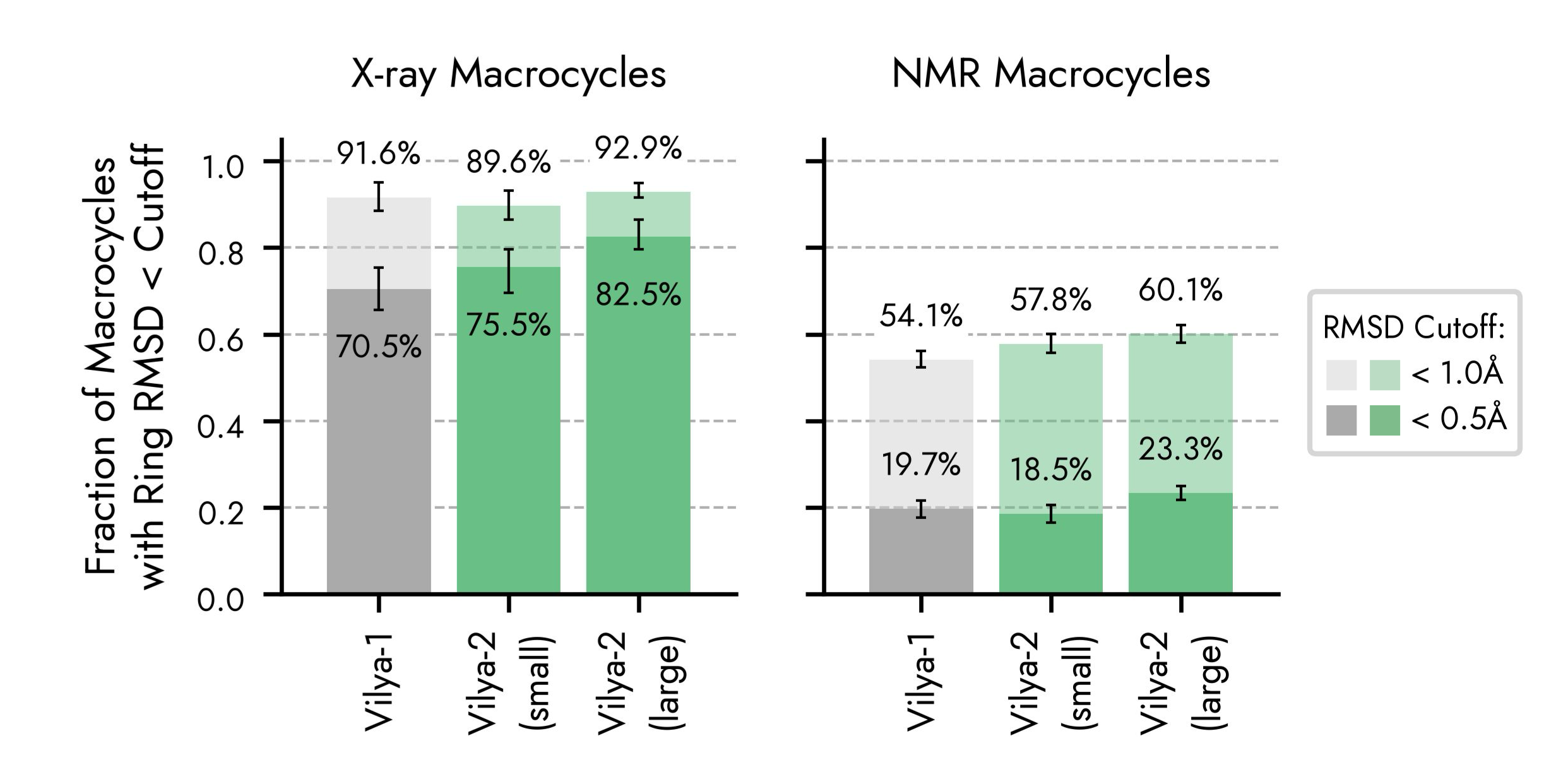}
  \caption{Conformer sampling success rates on a rigorous benchmark of X-ray characterized macrocycles. Both Vilya-2 variants exhibit superior structural prediction performance compared to Vilya-1, with particularly pronounced gains in the high-accuracy regime (dark bars; Ring RMSD $<$ 0.5\AA) compared to standard accuracy thresholds (light bars).}
  \label{fig:fig8}
\end{figure}

\paragraph{Vilya-2 is state-of-the-art at peptide conformer generation and property prediction.} Building on the architectural improvements described in the following section, we deploy the Vilya-2 conformer generator at two model scales, small and large. Both outperform Vilya-1 on the two benchmarks introduced in \citep{vilya1} (Figure \ref{fig:fig8}). On the X-ray macrocycle benchmark, the gains are most pronounced in the high-accuracy regime (ring RMSD $<$ 0.5\AA), where success rate increases by 5.1 and 12.1\% for the small and large models, respectively. Gains are also clear on the harder NMR benchmark, whose macrocycles are substantially larger: success rate (ring RMSD $<$ 1\AA) improves by 3.5 and 5.9\% for the small and large models, respectively. 

\begin{figure}[htbp]
  \centering
  \includegraphics[width=0.7\linewidth]{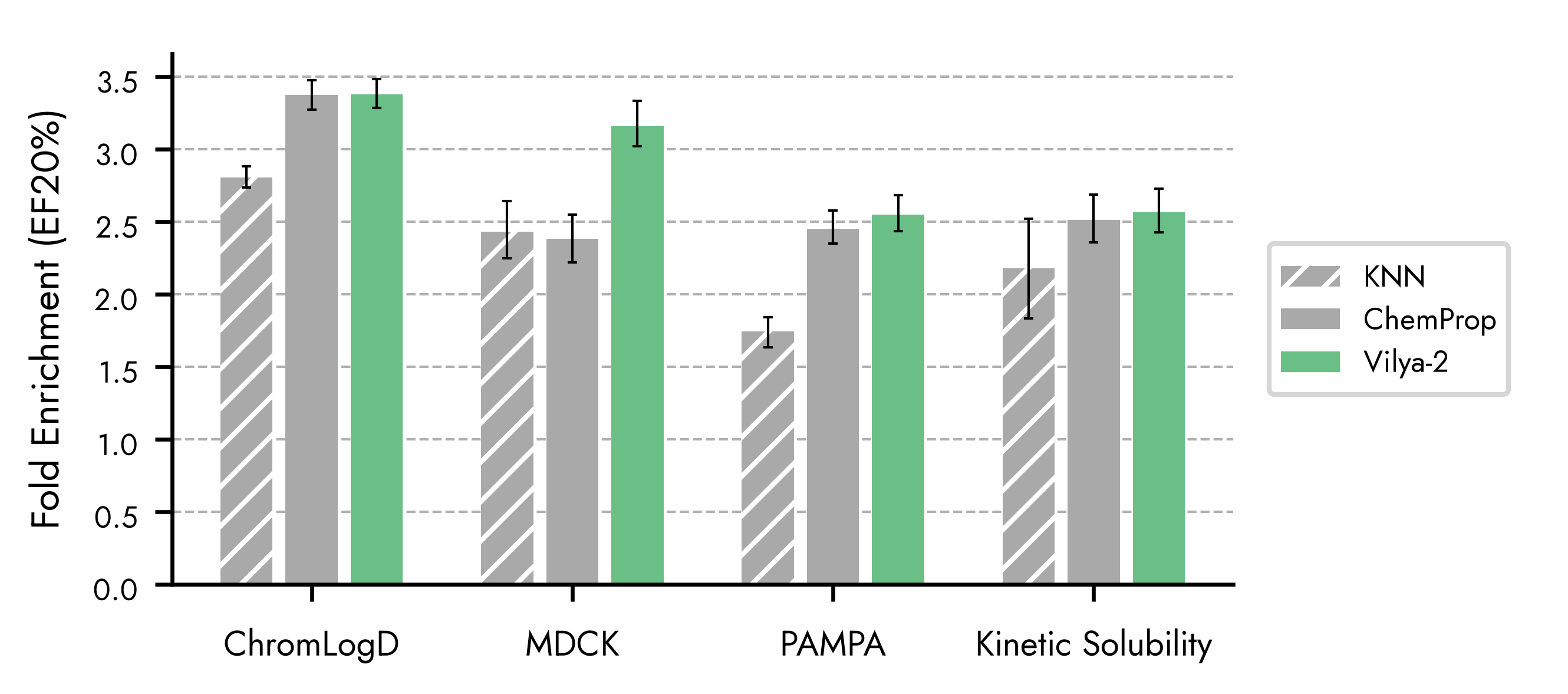}
  \caption{Vilya-2 property prediction results. Fold enrichment of favorable compounds in the top 20\% of predictions (EF20\%) for the Vilya-2 conformer generator fine-tuned as a property-prediction head (green), compared with two ligand-only baselines---K-nearest neighbors (KNN) on ECFP fingerprints (hatched) and ChemProp (gray)---on chromatographic logD (ChromLogD), MDCK permeability, PAMPA permeability, and kinetic solubility. Vilya-2 matches or exceeds both baselines on all four properties, with the largest margin on the MDCK permeability assay. For reference, the theoretical maximum enrichment on these datasets are 3.4, 4.9, 3.2, and 4.9, respectively}
  \label{fig:fig9}
\end{figure}

Following Vilya-1 \cite{vilya1}, we fine-tune the Vilya-2 conformer generator on internal macrocyclic data to predict four molecular properties related to bioavailability and exposure (Figure \ref{fig:fig9}). The largest gap in these ligand-only classes of models can be seen in Madin-Darby Canine Kidney cells (MDCK) permeability assay \cite{irvine1999mdck}, where access to a 3D conformation ensemble is beneficial to the model's enrichment of permeable compounds.

\begin{figure}[htbp]
  \centering
  \includegraphics[width=1.0\linewidth]{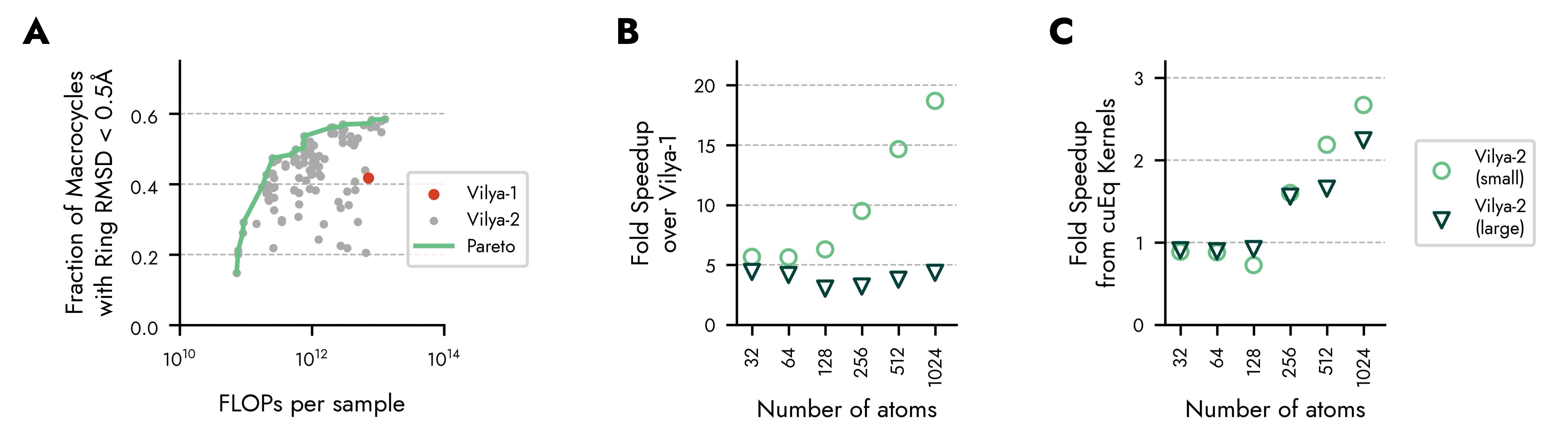}
  \caption{Comprehensive architectural optimization yields a highly efficient and accurate conformer generator. A) Empirical Pareto frontier mapping high-resolution structural accuracy against computational cost. The fraction of macrocycles achieving $<$ 0.5\AA~Ring RMSD is plotted as a function of inference FLOPs per sample. The Pareto front (green line) is derived from a systematic architectural sweep of 112 trained networks (gray dots) with varying hyperparameters, illustrating the scaling dynamics of the core architecture. B) Inference acceleration of the optimized Vilya-2 variants relative to the first-generation Vilya-1 baseline. Both the streamlined, high-throughput model (Vilya-2 small; light green circles) and the high-capacity model (Vilya-2 large; dark green triangles) execute considerably faster than Vilya-1 across diverse molecular sizes ranging from 32 to 1024 atoms, achieving up to a 19-fold speedup. C) Inference speedup from specialized cuEquivariance (cuEq) kernels for the same two variants, plotted against molecule size. The kernels provide little benefit at small atom counts but deliver increasing acceleration as system size grows, reaching roughly 2.5-fold at 1024 atoms---the regime relevant to folding larger miniproteins and modeling interfaces.}
  
  \label{fig:fig10}
\end{figure}

\paragraph{Vilya-2 is accelerated by architectural changes and specialized CUDA kernels.} Dropping triangle attention and analytically computing input gradients greatly improve both training and inference time relative to the Vilya-1 model. Enabled by this speed up, we trained and evaluated 112 candidate networks varying in parameter size and layer composition for conformer generation. We plot the empirical Pareto frontier in Figure \ref{fig:fig10}A. Analyzing the frontier of these models revealed distinct scaling dynamics. Initial increases in computational cost (FLOPs per sample) yielded rapid, substantial gains in high-resolution structural accuracy, measured by the fraction of macrocycles achieving $<$ 0.5\AA~Ring RMSD. Scaling further gives modest but diminishing returns at conformer generation, indicating a ceiling of performance limited by available training data and the intrinsic complexity of the benchmark.

Both small and large models show large speed gains relative to the previous Vilya-1 model: nearly $5\times$ throughput on smaller molecules and even larger gains for the smaller model at higher atom counts (Figure \ref{fig:fig10}B). The speed gains in Vilya-2 are partially driven by new, specialized kernels for two core operations---attention with pair bias and triangle multiplication---both provided by NVIDIA's cuEquivariance library \citep{cueq2025}. These improvements are largest at high atom counts (Figure \ref{fig:fig10}C), enabling the folding of larger miniproteins and allowing Vilya-2 to operate at the scales required for interface modeling. Together with compilation, these kernels reduce GPU memory usage during training by 42\%, enabling training with larger batches.

\section{Conclusion}
In this paper, we introduce Vilya-2, a neural network designed to model conformational ensembles and interfaces. Vilya-2 is unprecedented in its generality---showing superior performance relative to existing networks on not only small molecules but on challenging protein-peptide interfaces. It generalizes well to a variety of novel settings: to cross-docking, to protein-ligand complexes unseen in its training set, and to folding designed miniproteins without any notion of protein sequence or sequence-derived information. Vilya-2 is particularly good at ranking, and provides well-calibrated estimates of when it has made accurate predictions.

Vilya-2 goes beyond structure as a broadly applicable foundation model that can be fine-tuned for downstream tasks: in particular, it can learn to more accurately predict affinity measurements taken from real drug discovery campaigns than existing architectures. It can also be fine-tuned to predict properties beyond binding, particularly those properties relevant to designing orally bioavailable drugs. Vilya-2 represents a paradigm shift in molecular modeling: away from large co-folding networks and toward uniform chemical representations and fully-atomic context. This shift unlocks previously unseen generality in modeling arbitrary protein-ligand interactions, and forms the backbone of a broader effort to design broadly applicable, highly diverse peptide therapeutics.

\section{Contributors}

Pascal Sturmfels, Naozumi Hiranuma, Milad Salem, Benjamin D. Sellers, Stephen Rettie, CJ San Felipe, Chase~A.~P.~Wood, Jeffrey K. Holden, Adam P. Moyer, Patrick J. Salveson*, Ivan Anishchanka*

*Correspondence to: \href{mailto:patrick@vilya.ai}{patrick@vilya.ai}, \href{mailto:ivan@vilya.ai}{ivan@vilya.ai}.

\printbibliography

\appendix

\setcounter{figure}{0}
\renewcommand{\thefigure}{S\arabic{figure}}

\setcounter{table}{0}
\renewcommand{\thetable}{S\arabic{table}}

\section{Supplement}

\begin{table}[ht]
\centering
\caption{Ring RMSD of Vilya-2 predictions for the 21 disulfide-stapled miniproteins from three studies \citep{bhardwaj2016accurate,cabalteja2019heterogeneous,cabalteja2022heterogeneous}. For each miniprotein (PDB ID), the value is the lowest ring RMSD (Å) between any of the n = 4,000 generated Vilya-2 conformations and any member of the reference NMR ensemble.}
\label{tab:tabs1}
\begin{tabular}{lc|lc|lc|lc}
\toprule
PDB ID & Ring RMSD & PDB ID & Ring RMSD & PDB ID & Ring RMSD & PDB ID & Ring RMSD \\
\midrule
2ND2 & 0.87 & 5KWP & 0.78 & 5KX2 & 1.16 & 7RAP & 1.49 \\
2ND3 & 1.94 & 5KWX & 1.23 & 6E5H & 0.76 & 7TV5 & 0.75 \\
5JHI & 1.05 & 5KWZ & 0.41 & 6E5I & 0.58 & 7TV6 & 1.72 \\
5JI4 & 4.21 & 5KX0 & 0.50 & 6E5J & 1.36 & 7TV7 & 2.14 \\
5KVN & 0.86 & 5KX1 & 0.39 & 6E5K & 1.04 & 7TV8 & 2.10 \\
5KWO & 0.65 &      &      &      &      &      &      \\
\bottomrule
\end{tabular}
\end{table}

\begin{figure}[htbp]
  \centering
  \includegraphics[width=0.4\linewidth]{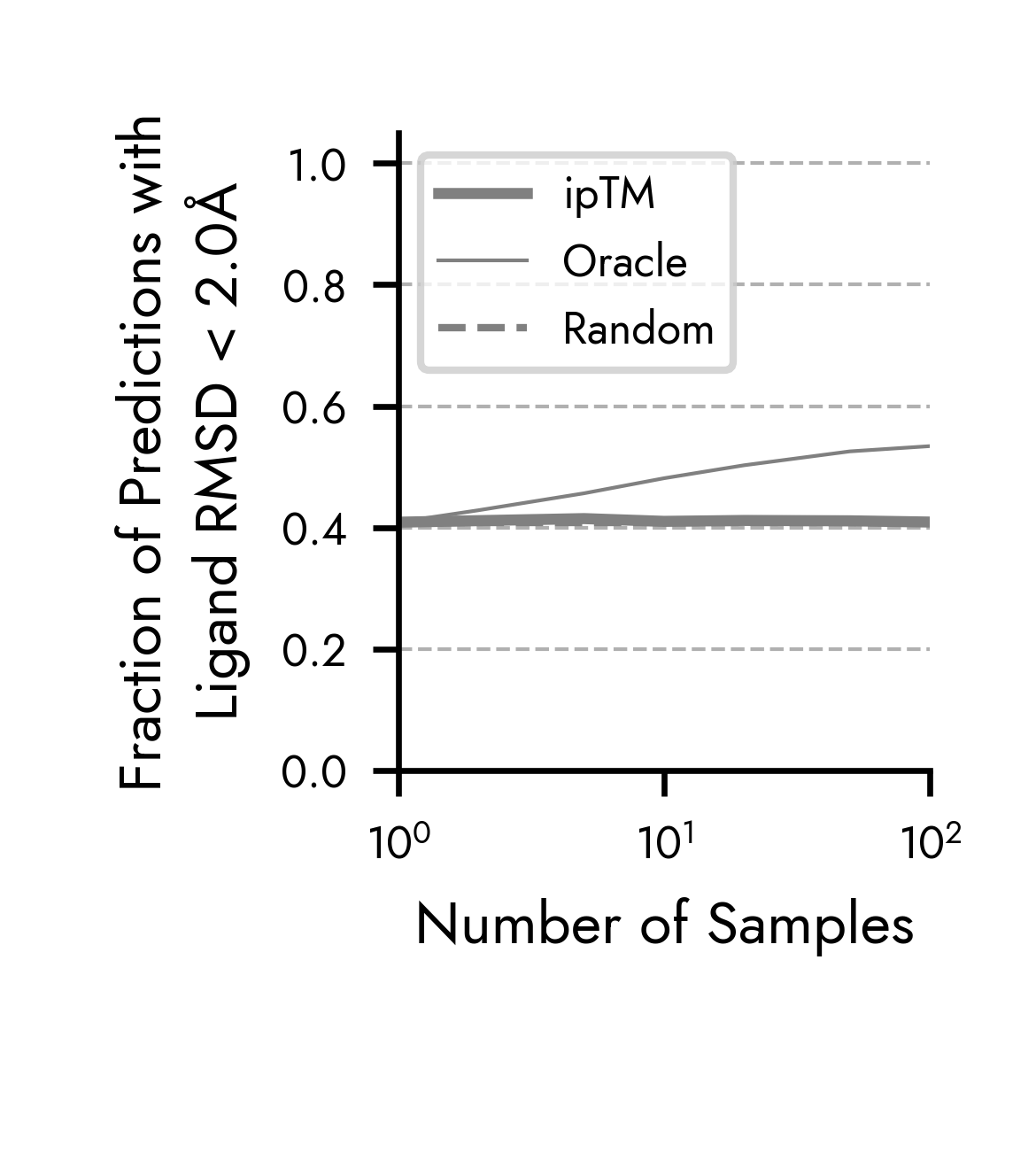}
  \caption{ Boltz-2 performance on the Runs N' Poses benchmark. Providing Boltz-2 with receptor conformation from the ground-truth co-crystal structure as an inference-time template (solid line), also shown on Figure \ref{fig:fig1}D only marginally increases success rates compared to the default settings (dashed line). }
  \label{fig:fig_s1}
\end{figure}

\begin{figure}[htbp]
  \centering
  \includegraphics[width=0.6\linewidth]{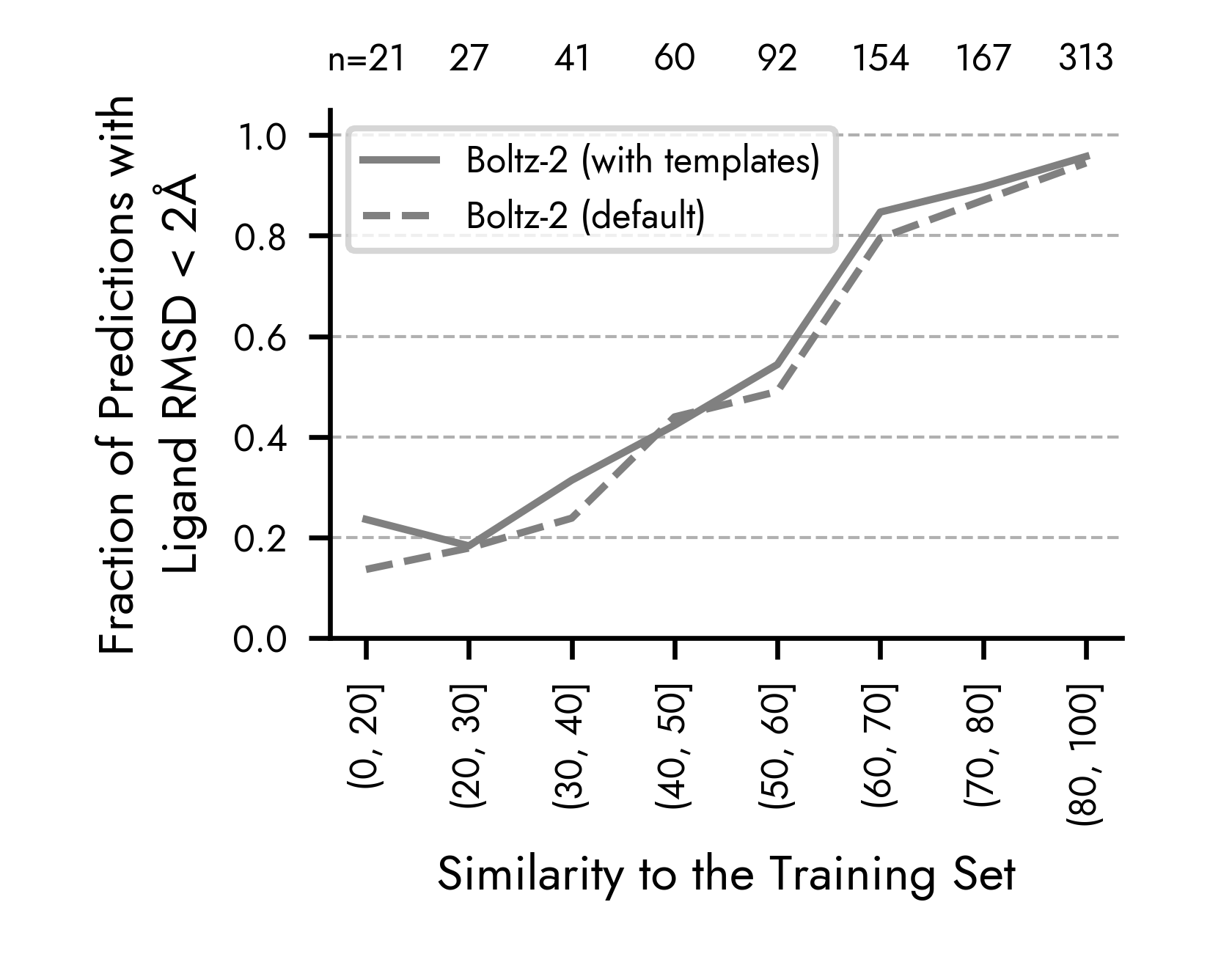}
  \caption{ Boltz-2 does not benefit from increased sampling on protein–peptide complexes. Increased sampling combined with ipTM-based ranking does not raise Boltz-2's success rate at protein–peptide complex prediction: the Random and ipTM lines overlap, indicating a lack of ranking power. }
  \label{fig:fig_s2}
\end{figure}

\begin{figure}[htbp]
  \centering
  \includegraphics[width=1.0\linewidth]{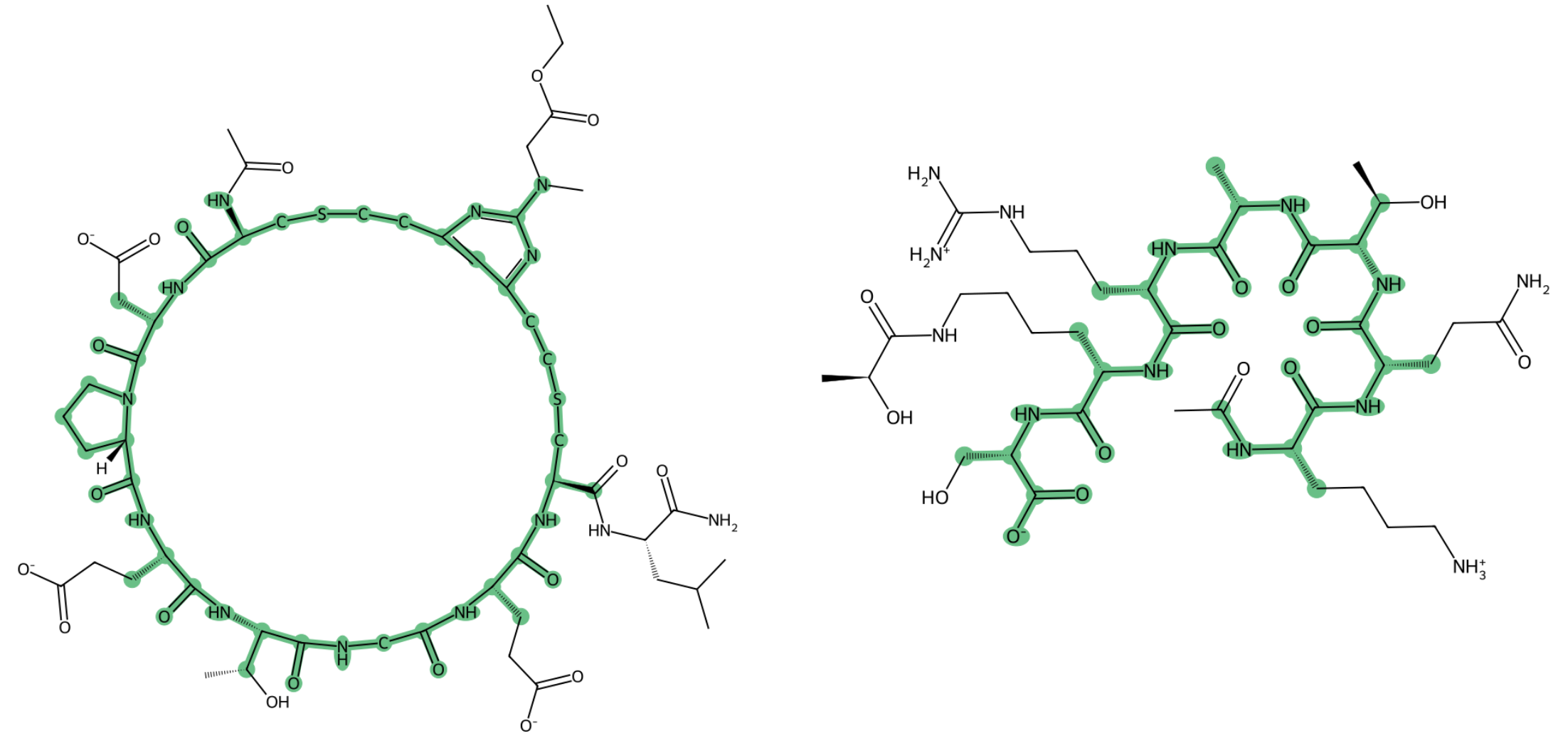}
  \caption{ Examples of the set of backbone atoms. The atoms in green are used to compute backbone RMSD on the peptide benchmark throughout the paper. Code associated with computing the backbone set of atoms is available in our public benchmarking code. }
  \label{fig:fig_s3}
\end{figure}

\begin{figure}[htbp]
  \centering
  \includegraphics[width=0.6\linewidth]{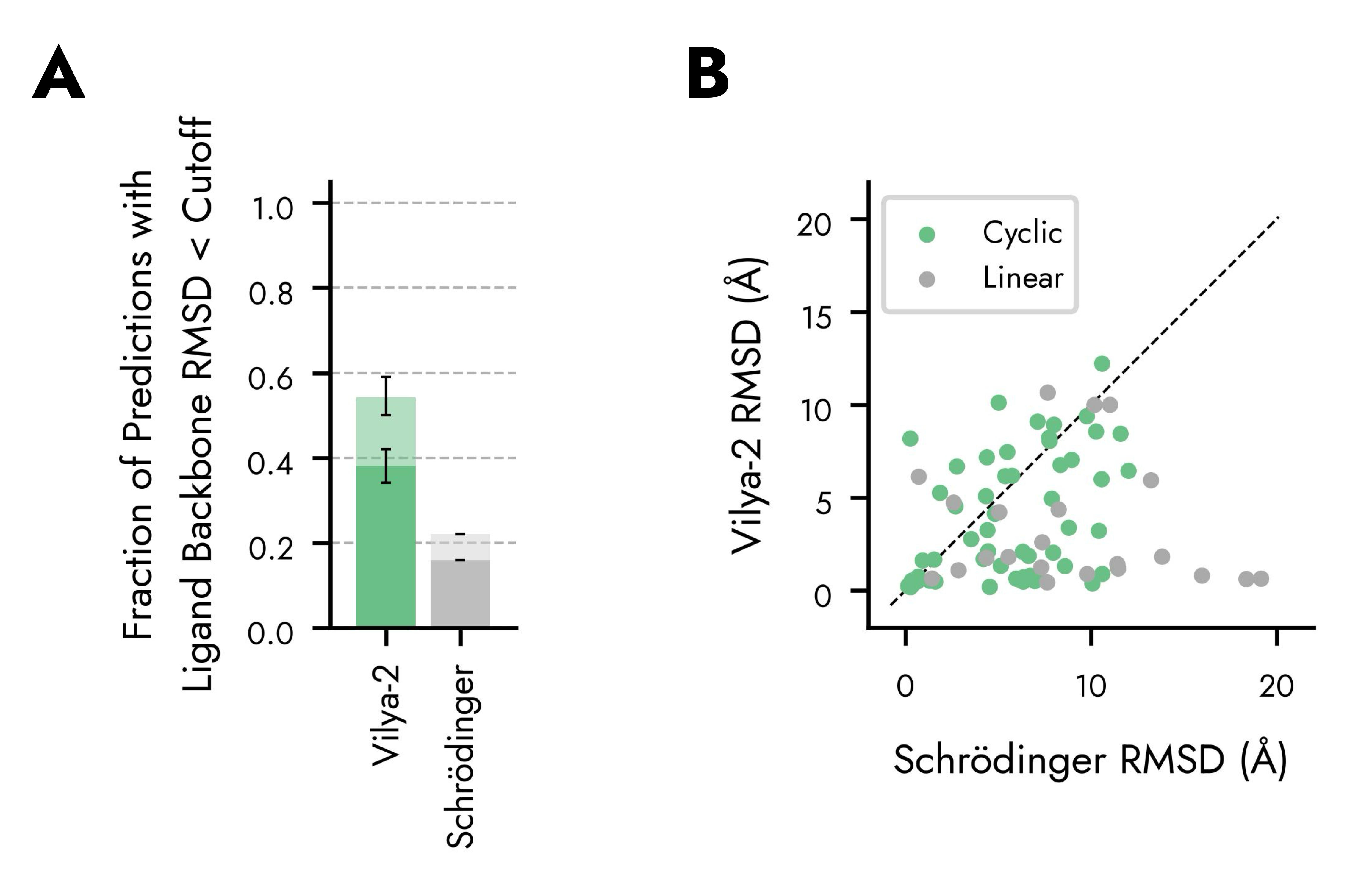}
  \caption{ Head-to-head comparison of Vilya-2 and Schrödinger on the Riptides benchmark. A) Fraction of predictions with ligand backbone RMSD below 2\AA~(light bars) and 1\AA~(dark bars) for Vilya-2 and Schrödinger (MacroDock for macrocycles, Glide for linear peptides). B) Per-target ligand backbone RMSD (\AA) for Vilya-2 (y-axis) versus Schr{\"o}dinger (x-axis), colored by peptide topology (cyclic, green; linear, gray). Points below the diagonal represent targets for which Vilya-2 produces a more accurate pose. The comparison in both panels is restricted to the 82 of 88 complexes for which Schrödinger docking succeeded. }
  \label{fig:fig_s4}
\end{figure}

\end{document}